\documentclass{article} 
\usepackage{iclr2025,times}

\usepackage{amsmath,amsfonts,bm}

\def\eqref#1{equation~\ref{#1}}

\def\1{\bm{1}}

\DeclareMathAlphabet{\mathsfit}{\encodingdefault}{\sfdefault}{m}{sl}
\SetMathAlphabet{\mathsfit}{bold}{\encodingdefault}{\sfdefault}{bx}{n}

\usepackage{hyperref}
\usepackage{url}

\usepackage{graphicx}
\usepackage{float}
\usepackage{subcaption}

\title{Do Not Overestimate Black-box Attacks}

\author{Han Wu \\ 
Department of Physics and Astronomy\\
University of Southampton\\
\texttt{\{han.wu\}@soton.ac.uk} \\
\And
Sareh Rowlands \& Johan Wahlstr\"om \\
Department of Computer Science \\
University of Exeter \\
\texttt{\{s.rowlands, j.wahlstrom\}@exeter.ac.uk} \\
}

\iclrfinalcopy 
\begin{document}

\maketitle

\begin{abstract}
As cloud computing becomes pervasive, deep learning models are deployed on cloud servers and then provided as APIs to end users. However, black-box adversarial attacks can fool image classification models without access to model structure and weights. Recent studies have reported attack success rates of over 95\% with fewer than 1,000 queries. Then the question arises: whether black-box attacks have become a real threat against cloud APIs? To shed some light on this, our research indicates that black-box attacks against cloud APIs are not as effective as proposed in research papers due to several common mistakes that overestimate the efficiency of black-box attacks. To avoid similar mistakes, we conduct black-box attacks directly on cloud APIs rather than local models.
\end{abstract}




\section{Introduction}


Image classification models are widely used in real-world applications , often achieving top-5 accuracy exceeding 90\%. Cloud-based image classification services, such as Google Cloud Vision, offer pre-trained models as APIs, allowing users to classify images by sending requests to cloud servers. This is useful for IoT devices that lack the computational power to run deep learning models locally.

\begin{figure*}[tbph]
    \centering
    \includegraphics[width=\linewidth]{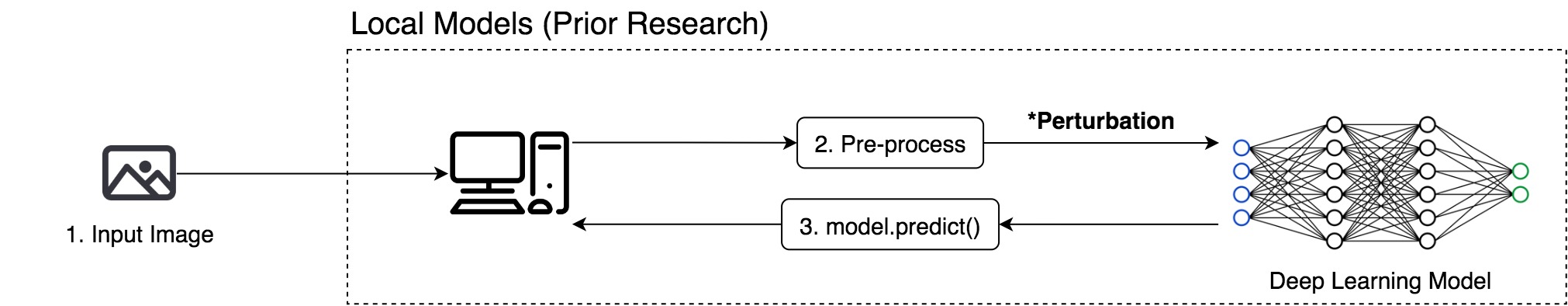}
    \caption{Most prior research tests black-box attacks on local models, where the adversarial perturbation is applied after pre-processing and just before the input is fed into deep learning models, assuming access to the input of a black-box model.}
    \label{fig:local}
\end{figure*}

\begin{figure*}[btph]
    \centering
    \includegraphics[width=\linewidth]{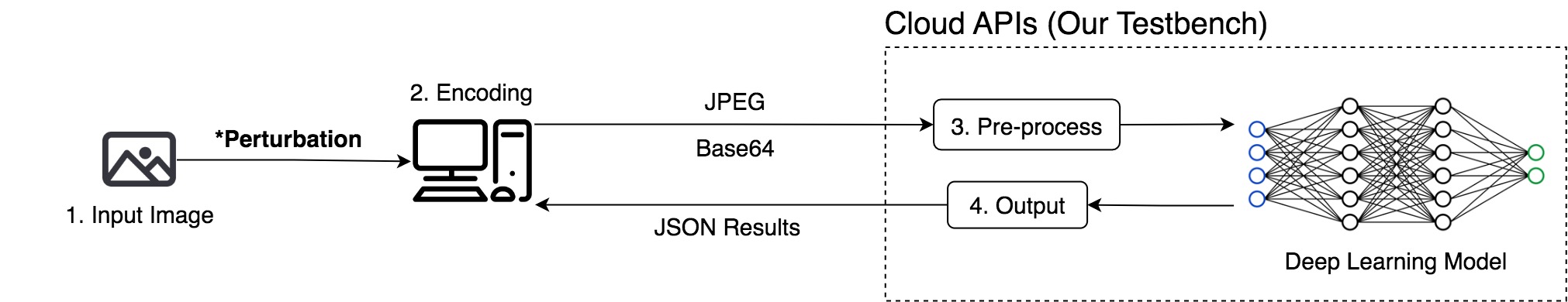}
    \caption{We initiate black-box attacks directly against cloud APIs, applying the adversarial perturbation before image encoding and pre-processing. This approach assumes no access to the internal workflow of cloud-based black-box models.}
    \label{fig:cloudapi}
\end{figure*}

However, image classification cloud services are vulnerable to black-box adversarial attacks, which generate imperceptible perturbations on input images to mislead classification models. Although prior research has shown that black-box attacks can achieve higher than 95\% success rates with only 1,000 queries without access to model structure and weights \citep{bhambri2019survey}, most research generates adversarial images \textbf{offline} on local models (see Fig.\ref{fig:local}) rather than \textbf{online} on cloud APIs (see Fig. \ref{fig:cloudapi}), thereby inadvertently exploits information that is unavailable for cloud-based black-box models. The actual efficiency of online black-box attacks against cloud services remains unclear.

Black-box attacks generate adversarial images by sending queries to the target model. Implementing \textbf{online} black-box attacks is more challenging because cloud APIs generally have slower response times compared to \textbf{offline} attacks against local models. While local models with GPU acceleration can respond to more than 100 queries per second, the typical response time from an API server is 0.5 - 2s per query. Online black-box attacks pose a limited practical threat because generating multiple adversarial images could take several hours. Therefore, online attacks must be both time-efficient and can achieve high success rates. However, previous research often underestimates the time consumption and overestimates the attack success rate.

\section{Methodology}
\label{common_errors}

\subsection{Problem Formulation}

Given an input image $x$ and the true label $y$, the objective of the adversary is to add a small perturbation $\delta$ to the original image, and generate an adversarial image $x' = x\ +\ \delta$ that can fool a black-box image classifier $C(x)$, such that $C(x') \neq C(x)$. Typically, the perturbation $\delta$ is bounded by the $l_2$ or $l_\infty$ norm with user-defined constants $\epsilon$ \citep{bhambri2019survey}.

The adversary does not know the model structure and weights. Further, the attacker has limited access to model outputs for cloud-based black-box models. For example, in the partial-information setting, the adversary only has access to the prediction probabilities of the top $k$ classes $\{y_1, ..., y_k\}$. In the label-only setting, the adversary can only access the prediction label without any knowledge of prediction probabilities \citep{ilyas2018black}.

\subsection{Common Mistakes}

We observed that some previous research made similar mistakes in the query process, which provided their attacks with an unfair advantage. This advantage led these methods to outperform state-of-the-art black-box attacks, but it was based on the assumption of accessing information that is not available in black-box attacks. These issues are present in several widely-used black-box attacks, including Bandits Attack \citep{ilyas2018black, ilyas2018prior}, SimBA Attack \citep{guo2019simple}, Parsimonious Attack \citep{ moon2019parsimonious}, Square Attack \citep{andriushchenko2020square} and some recently published research \citep{liu2024difattack, park2024hard, ran2025black}. It is important to raise awareness within the research community and prevent similar mistakes in future publications.
\subsubsection{Image Encoding}

 In real-world scenarios, images are encoded before being sent to cloud services to reduce the amount of data transmitted and save bandwidth. However, prior research, as mentioned above, assumes that perturbations can be added directly to the raw input of deep neural networks (see Fig. \ref{fig:local}). 

 Cloud services such as Google Cloud Vision and Imagga accept raw binary and base64-encoded JPEG images as input. Since JPEG compression is lossy, it may discard some of the perturbations during encoding, thereby reducing the success rate of attacks \cite{dziugaite2016study}. Therefore, evaluating black-box attacks on local models without considering image encoding could lead to an overestimation of their effectiveness against real-world cloud APIs.

\subsubsection{Image Pre-processing}

Papers listed above apply perturbations after image resizing, implicitly assuming knowledge of the input shape of the black-box model. Moreover, note that original input images are typically larger than the model input shape. Resizing high-resolution images to a lower resolution reduces the sampling space, thereby making it less computationally intensive to generate perturbations. 

Besides, image classification cloud services do not accept images with invalid pixel values (pixel value $>$ 255 or $<$ 0). For example, the Bandits Attack does not clip the pixel value of adversarial images, and thus overestimate the attack success rate by sending invalid pixel values to the model.

\clearpage

\subsection{Possible Causes}

Most prior research tests their attacks on local models rather than on online models because it is both faster and less costly. Sending queries to real-world cloud APIs typically costs around \$1 for every 1,000 requests (for example, using Google Cloud Vision). In other words, an experiment attacking 1,000 images and maintaining a query budget of 1,000 queries per image would require 1,000,000 queries and cost \$1000. 

As a result, most prior research evaluates their attacks on local models and relies on themselves to restrain access to extra model information. However, either intentionally or unintentionally, they exploit extra information that enhances their attacks for the following reasons:

\begin{itemize}
    \item The PyTorch prediction function only accepts input images $x$ as an array with the same shape. Thus, it is tempting to resize all input images to match the model's required input size, thereby exploiting extra information about the model input shape.
    \item The PyTorch prediction function accepts input images as floating-point numbers and does not produce an error even if the input image $x$ contains negative values. Consequently, prior research, without considering image encoding, can sent invalid pixel values to the model.
\end{itemize}

\subsection{Solutions}

To avoid these common mistakes, we designed an open-source image classification cloud service, named DeepAPI (see Appendix \ref{deepapi}), to ensure adversarial perturbations are generated and applied before image encoding and pre-processing.

Additionally, we provide an open-source Black-box Adversarial Toolbox that demonstrates how to conduct online black-box attacks against cloud APIs (see Appendix \ref{bat}), with a focus on practical considerations in real-world scenarios.



\section{Experimental Results}

\subsection{Black-box Adversarial Attacks}
\label{black_box}

Black-box attacks aim to deceive deep-learning models without having access to their internal structure or weights. We evaluated two common types of black-box attacks: Gradient Estimation and Local Search methods, which have been widely studied in the literature \citep{bhambri2019survey, wang2022black}.

\textbf{Local Search Methods}: The task of generating adversarial inputs can be approached as a problem of selecting what pixels to attack. Thus, we can use existing local search methods to search for combinations of pixels to be perturbed. One simple, yet effective, baseline attack that use this idea is the Simple Black-box Attack (SimBA) \citep{guo2019simple}. With SimBA, a vector is randomly sampled from a predefined orthonormal basis and then added or subtracted from the image. 

To improve sample efficiency, Andriushchenko et al. proposed the Square Attack \citep{andriushchenko2020square}. This attack initializes the perturbation using vertical stripes because CNNs are sensitive to high-frequency perturbations \citep{yin2019fourier}, and then generates square-shaped perturbations at random locations to deviate model predictions.

\textbf{Gradient Estimation Methods}: Inspired by white-box attacks that use gradients to generate adversarial perturbations \citep{GoodfellowSS14} \citep{madry2017towards}, gradient estimation methods estimate gradients through queries, and then use these estimated gradients to construct adversarial perturbations. To estimate gradients, Chen et al. used the finite-differences method to compute the directional derivative at a local point \citep{chen2017zoo}. To improve query efficiency, Ilyas et al. proposed a natural evolutionary strategy (NES) based method \citep{wierstra2014nes} to approximate gradients, and proved that the standard least-squares estimator is an optimal solution to the gradient-estimation problem \citep{ilyas2018black}. 

In our experiments, we evaluated the Bandits Attack, which further improved the classifier by using priors on the gradient distribution \citep{ilyas2018prior}, thereby exploiting the fact that the gradients at the current and previous steps are highly correlated.

\clearpage

\subsection{Attacking Local Model and Cloud APIs}

We evaluated three black-box attacks, SimBA, Square Attack, and Bandits Attack, using 1,000 images, each belonging to a unique class in ImageNet. For all attacks, we applied perturbations with a consistent strength of $\epsilon = 0.05$ under the $L_{\infty}$ norm\footnote{Our source code: \url{https://github.com/wuhanstudio/adversarial-classification/}.}. Our experimental results reveal that these attacks achieve significantly lower success rates when attacking cloud APIs.

\begin{figure*}[tbph]
\centering
\begin{subfigure}[t]{0.32\textwidth}
    \centering
    \includegraphics[width=\textwidth]{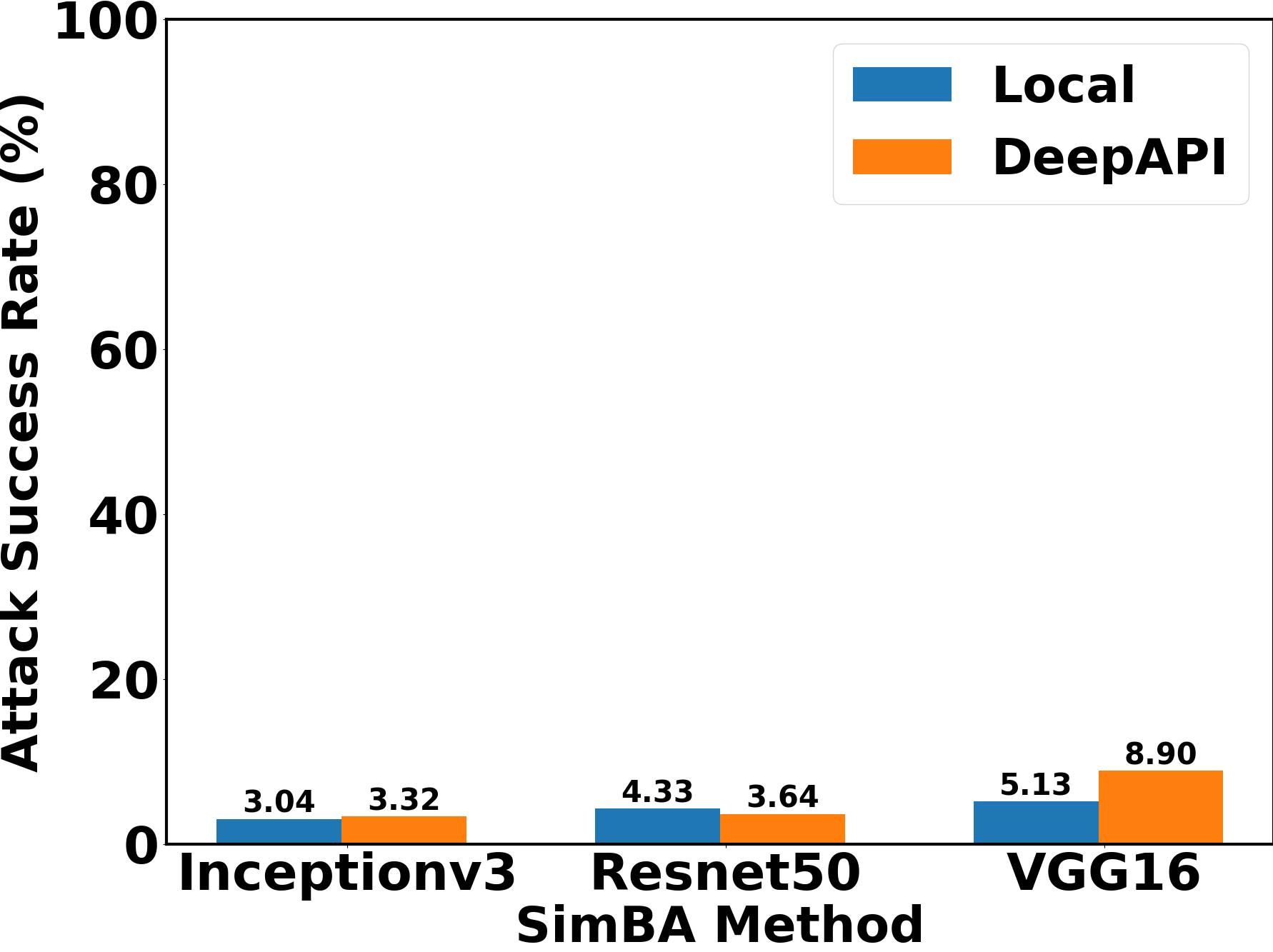}
    \caption{SimBA (Baseline)}
    \label{fig:simba_suc}
\end{subfigure}
\hfill
\begin{subfigure}[t]{0.32\textwidth}
    \centering
    \includegraphics[width=\textwidth]{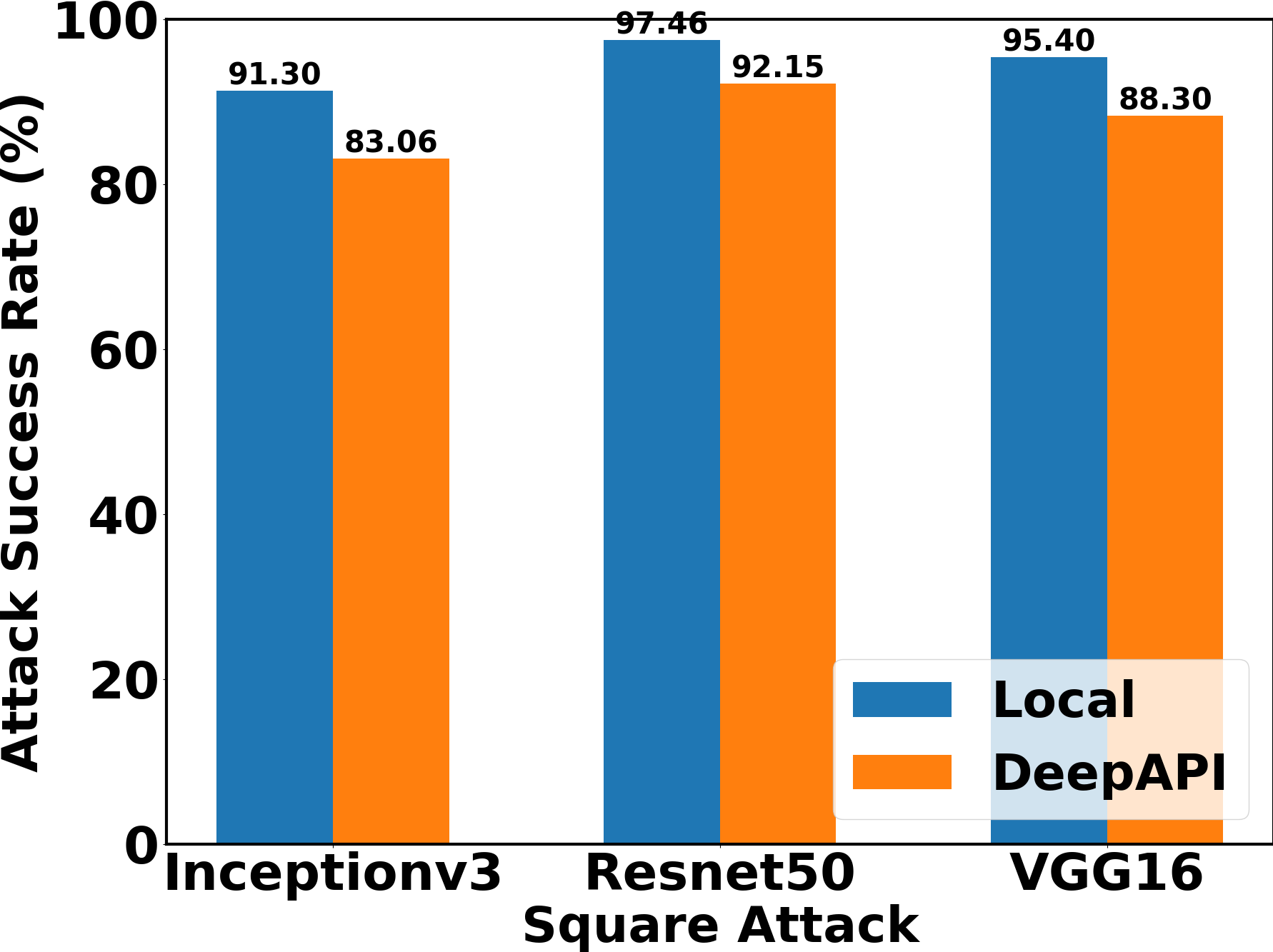}
    \caption{Square Attack (Local Search)}
    \label{fig:square_suc}
\end{subfigure}
\hfill
\begin{subfigure}[t]{0.32\textwidth}
    \centering
    \includegraphics[width=\textwidth]{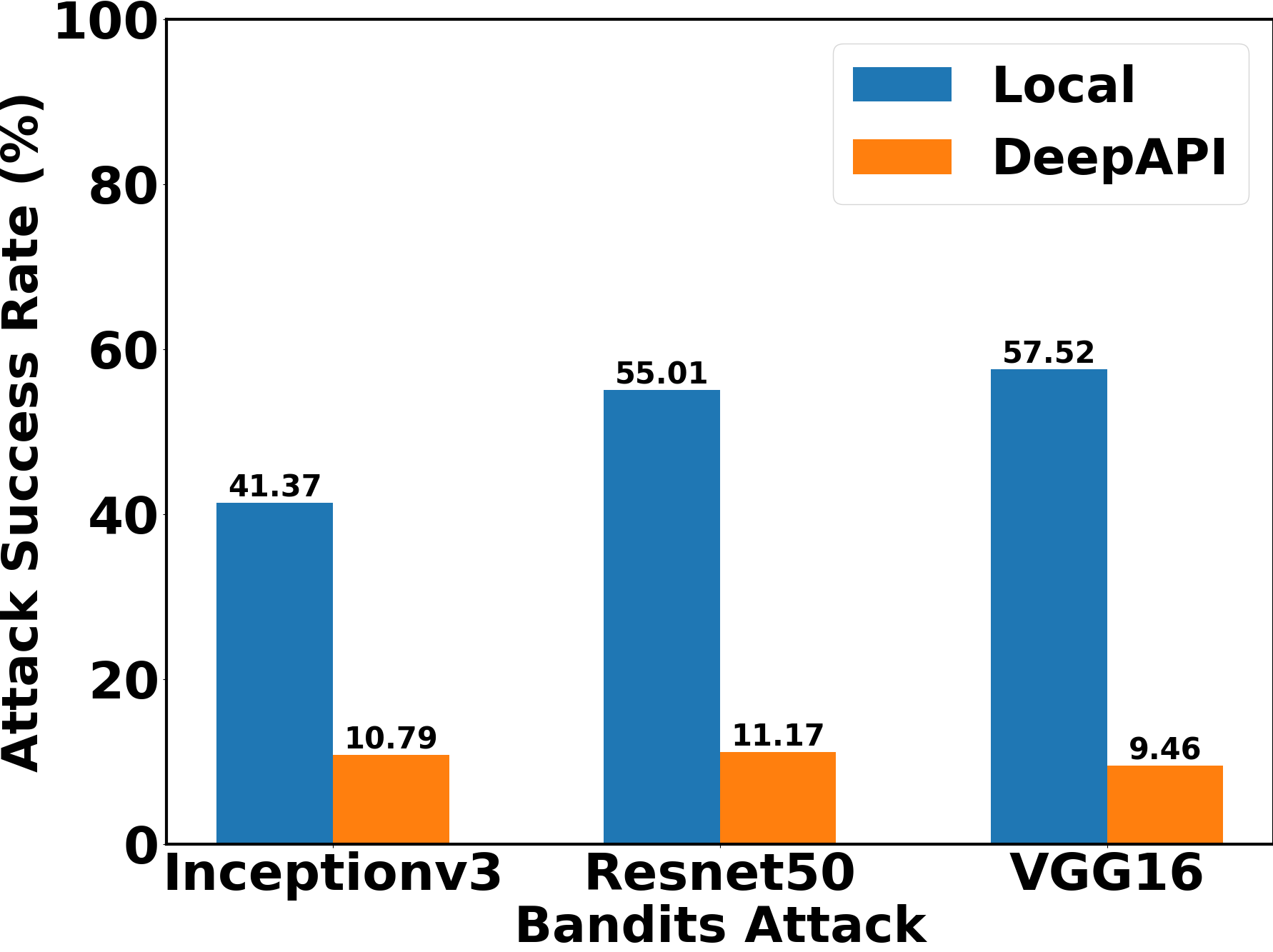}
    \caption{Bandits Attack (Gradients)}
    \label{fig:bandits_suc}
\end{subfigure}
\caption{The attack success rate of attacking local models and cloud APIs.}
\label{fig.suc}
\end{figure*}

\begin{figure*}[tbph]
\centering
\begin{subfigure}[t]{0.32\textwidth}
    \centering
    \includegraphics[width=\textwidth]{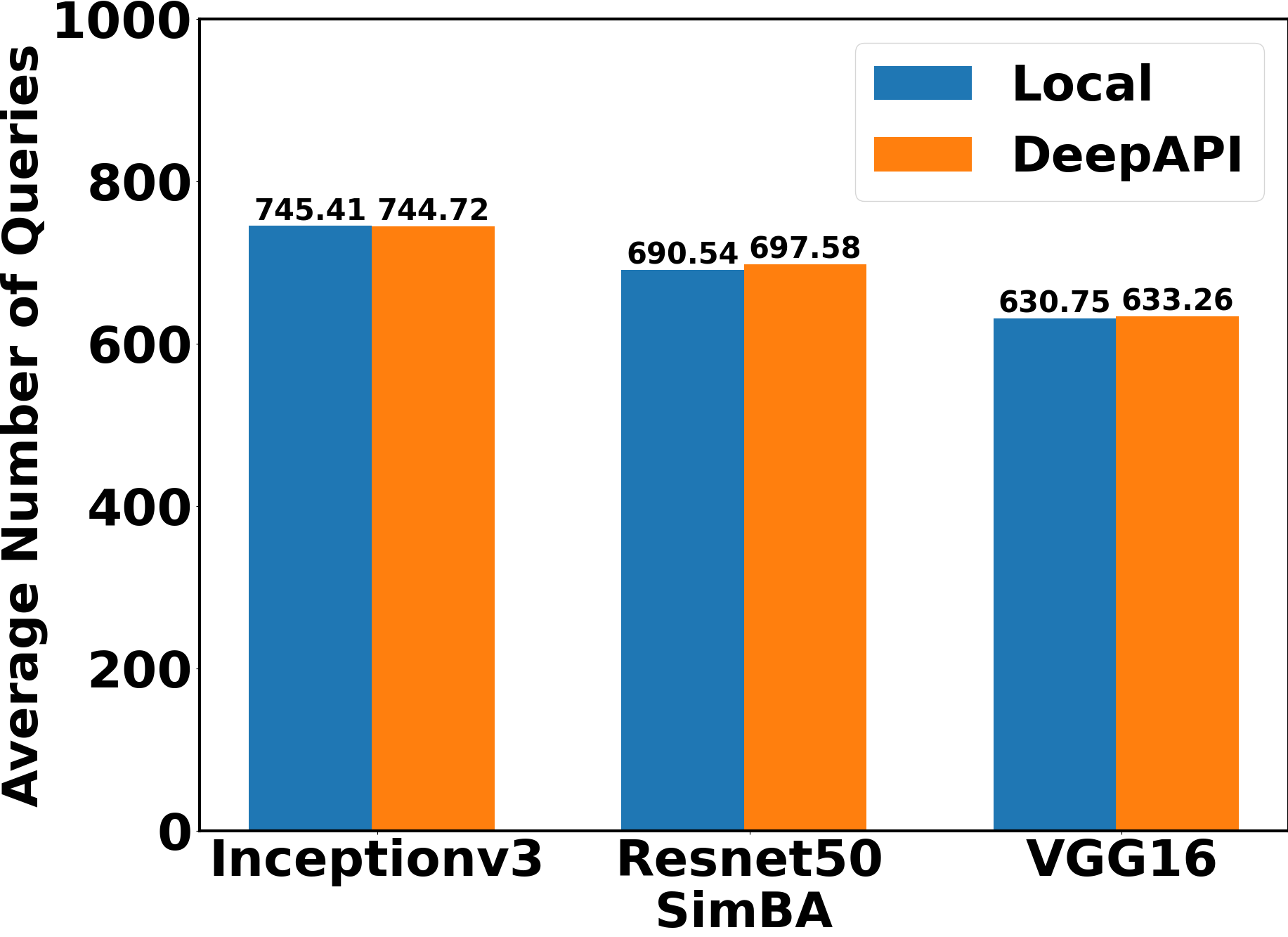}
    \caption{SimBA (Baseline)}
    \label{fig:simba_queries}
\end{subfigure}
\hfill
\begin{subfigure}[t]{0.32\textwidth}
    \centering
    \includegraphics[width=\textwidth]{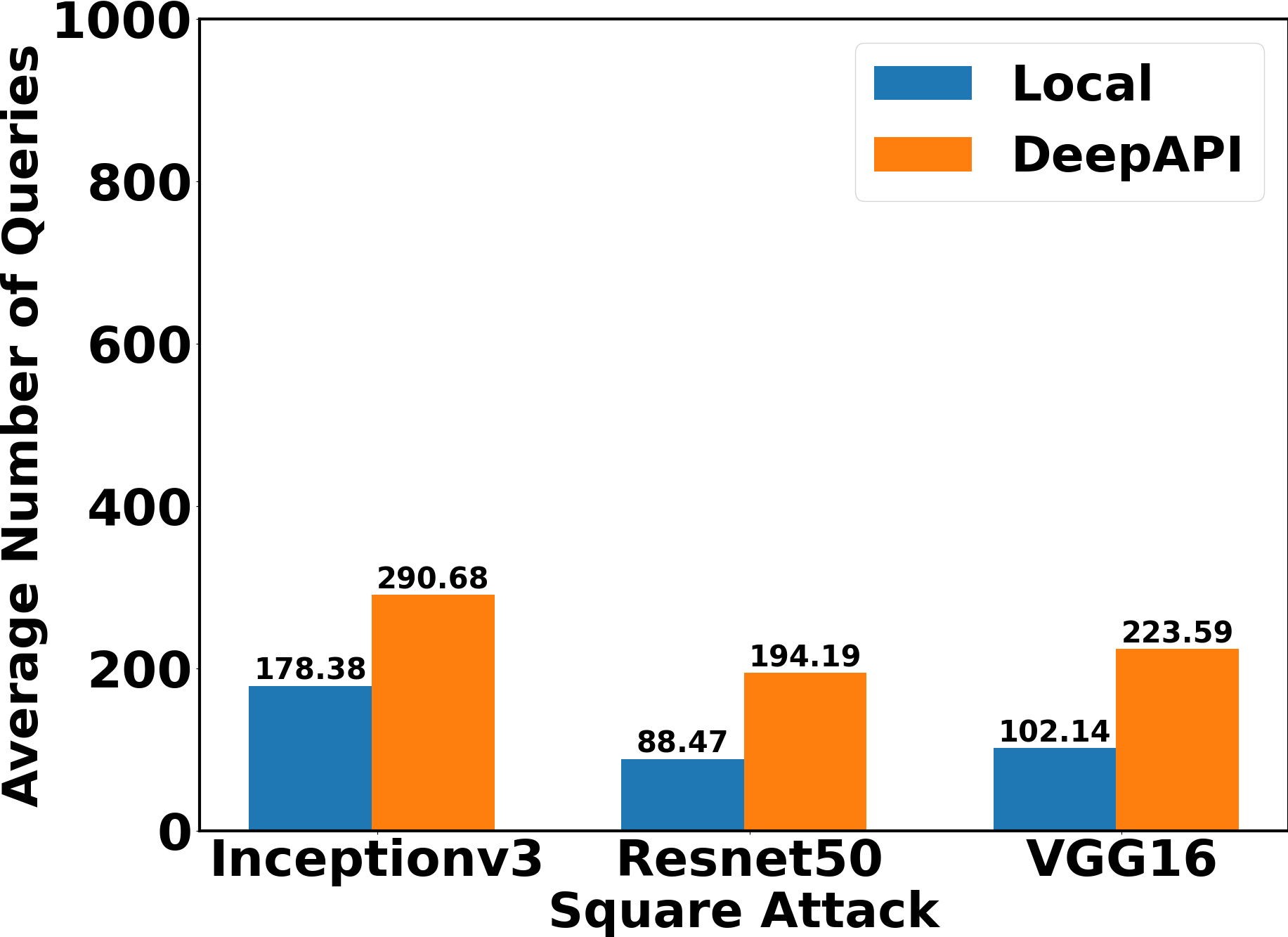}
    \caption{Square Attack (Local Search)}
    \label{fig:square_queries}
\end{subfigure}
\hfill
\begin{subfigure}[t]{0.32\textwidth}
    \centering
    \includegraphics[width=\textwidth]{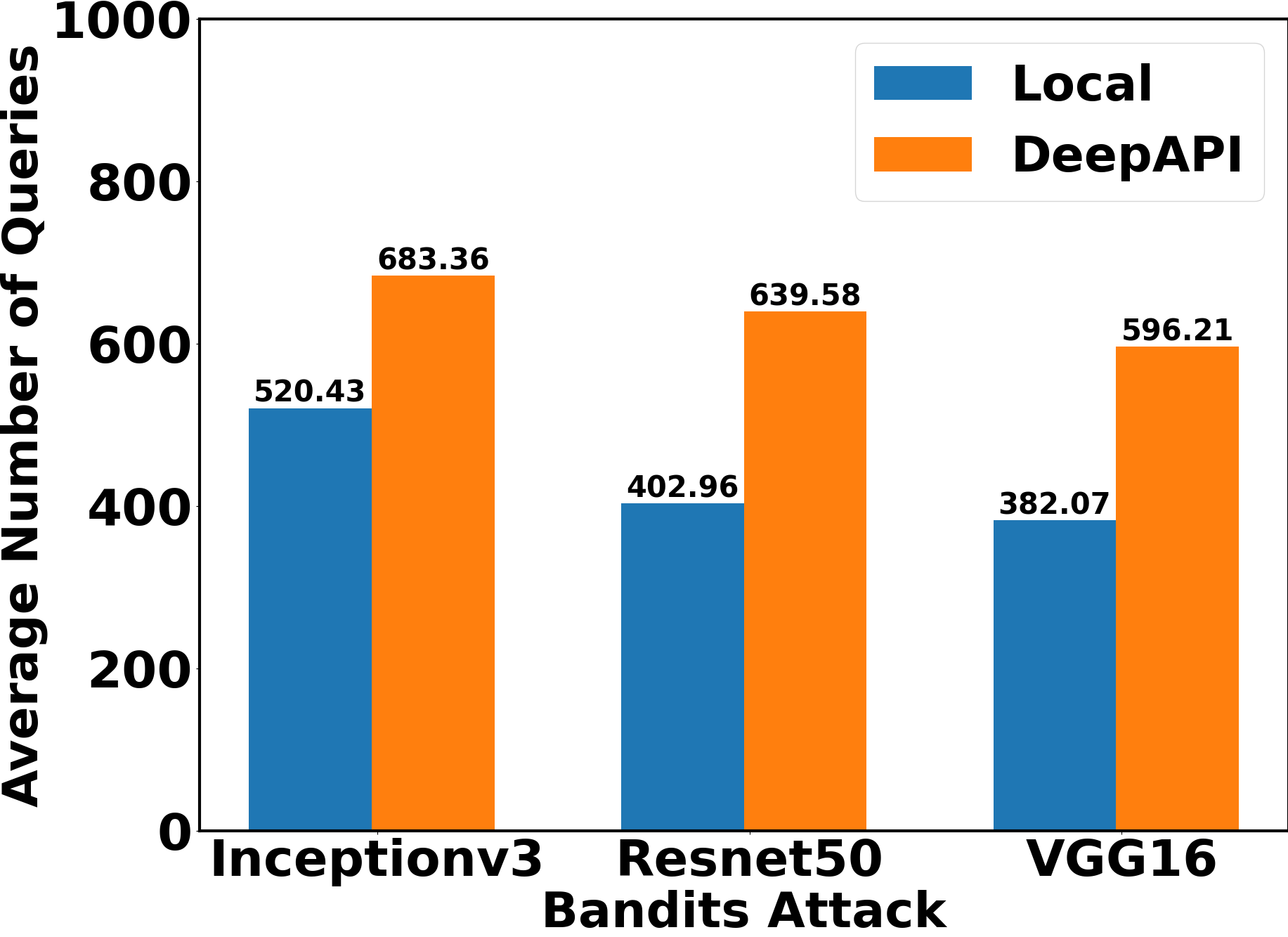}
    \caption{Bandits Attack (Gradients)}
    \label{fig:bandits_queries}
\end{subfigure}
\caption{The average number of queries of attacking local models and cloud APIs.}
\label{fig.queries}
\end{figure*}

The \textbf{SimBA Attack}, a baseline method, achieves comparable low attack success rates and requires similar number of queries for both local models and cloud APIs (see Figs. \ref{fig:simba_suc} and \ref{fig:simba_queries}). However, it is important to note that the success rate of SimBA is relatively low (approximately 5\%), and most attacks exhaust the full query budget (1,000 queries). 


\textbf{Square Attack}, a local search method, applies perturbations to high-resolution images when attacking cloud APIs, resulting in a  a lower success rate (Fig. \ref{fig:square_suc}) and requires more queries (Fig. \ref{fig:square_queries}). Due to the absence of image resizing, it is more challenging to find adversarial examples in a larger space, and thus the attack against cloud APIs is less effective \cite{guo2017countering}. 

\textbf{Bandits Attack}, a gradient estimation method, struggles to estimate gradients accurately before image resizing. Bilinear interpolation creates low-resolution images by subsampling from high-resolution inputs, resulting in zero gradients at unsampled points, which makes it difficult to produce valid estimates. As a result, the attack success rate against cloud APIs is significantly lower compared to attacks on local models (Figs. \ref{fig:bandits_suc} and \ref{fig:bandits_queries}).

\section{Conclusion}

This paper aims to investigate if black-box adversarial attacks have become a practical threat against image classification cloud services. We identify some common mistakes in prior research that leads to an overestimation of the efficiency of black-box attacks. 

Additionally, we contribute to the research community by open-sourcing our image classification cloud service, DeepAPI, and Black-box Adversarial Toolbox to facilitate future research on practical black-box attacks against cloud APIs.

\clearpage

\bibliography{iclr2025}
\bibliographystyle{iclr2025}

\clearpage

\appendix
\section{DeepAPI}
\label{deepapi}

\begin{figure*}[bthp]
    \centering
    \includegraphics[width=\linewidth]{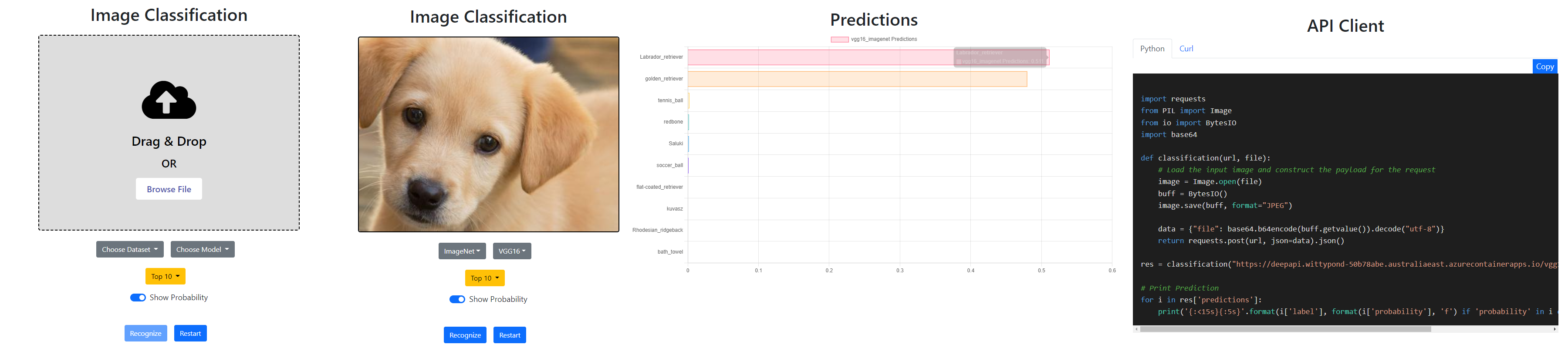}
    \caption{DeepAPI provides both web interface and APIs for research on black-box attacks.}
    \label{fig:deepapi}
\end{figure*}

To facilitate future research on black-box attacks that attack cloud APIs rather than local models, we designed DeepAPI \footnote{The source code of DeepAPI is available on Github: \url{https://github.com/wuhanstudio/DeepAPI/}.}, an open-source image classification cloud service (see Fig. \ref{fig:deepapi}) that supports:

\begin{itemize}
    \item The three most commonly evaluated classification models in existing research on black-box attacks: VGG16, ResNet50, and Inceptionv3 provided by Keras model zoo.
    \item Both soft labels (with probabilities) and hard labels (no probabilities) for label-only setting.
    \item Top $k$ predictions ($k \in \{1, 3, 5, 10\}$) for partial-information setting.
\end{itemize}

\section{Black-box Adversarial Toolbox}
\label{bat}

\begin{figure*}[btph]
    \centering
    \includegraphics[width=\linewidth]{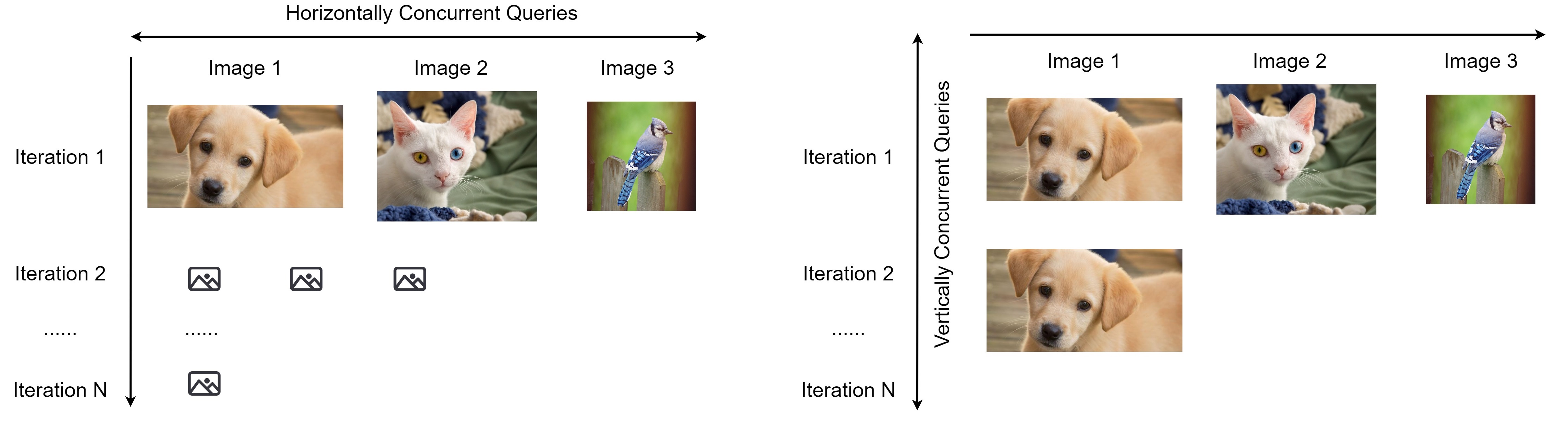}
    \caption{The difference between horizontal and vertical distribution.}
    \label{fig:distributability}
\end{figure*}

To demonstrate how to implement online black-box attacks against cloud APIs, we open-source the Black-box Adversarial Toolbox\footnote{The source code of the Black-box Adversarial Toolbox is available on GitHub: \url{https://github.com/wuhanstudio/blackbox-adversarial-toolbox}.}. To further enhance the practicality of existing black-box attacks, we propose horizontal and vertical distribution strategies (see Fig. \ref{fig:distributability}), inspired by the horizontal and vertical scaling of cloud resources \citep{Millnert2020}.

Horizontal Distribution concurrently sends queries for different images within the same iteration, thereby allowing the generation of multiple adversarial examples concurrently. This can be achieved without altering existing black-box attack methods. Since horizontal distribution does not require significant modifications to the original attack method, we can apply horizontal distribution by implementing a distributed query function that sends concurrent requests to cloud APIs.

Vertical Distribution, on the other hand, sends multiple concurrent queries for the same image, thereby accelerating the attack for that particular image. Existing black-box attack methods need to be redesigned to decouple the queries across iterations.

In summary, horizontal distribution achieves concurrent attacks against multiple images, while vertical distribution speeds up attacks on a single image.

\end{document}